# Urdu Dependency Parsing and Treebank Development: A Syntactic and Morphological Perspective

Nudrat Habib
Computer Science department
COMSATS University Islamabad,
Abbottabad Campus, Pakistan
nudrat@cuiatd.edu.pk

**Abstract:** Parsing is the process of analyzing a sentence's syntactic structure by breaking it down into its grammatical components. and is critical for various linguistic applications. Urdu is a low-resource, free word-order language and exhibits complex morphology. Literature suggests that dependency parsing is well-suited for such languages. Our approach begins with a basic feature model encompassing word location, head word identification, and dependency relations, followed by a more advanced model integrating part-of-speech (POS) tags and morphological attributes (e.g., suffixes, gender). We manually annotated a corpus of news articles of varying complexity. Using Maltparser and the NivreEager algorithm, we achieved a best-labeled accuracy (LA) of 70% and an unlabeled attachment score (UAS) of 84%, demonstrating the feasibility of dependency parsing for Urdu.

## 1. INTRODUCTION

Parsing is the process of structuring a linear representation in accordance with a given grammar [1]. Linear representation can range from a simple sentence to a complex computer program [2]. Parsing involves dividing a sentence into its grammatical components while identifying the relationships between these components [3]. The applications of parsers extend beyond linguistics into various fields.[4]. Parse trees are instrumental in examining the grammatical structure of sentences and serve as a crucial intermediary in semantic analysis and are also vital in question answering system[5]. Additionally, parsing techniques are employed in machine translation to resolve lexical ambiguities, as well as in information extraction [6], and information retrieval addressing challenges such as polysemy and synonymy [7].

Broadly there are two views of linguistic structure, phrase structure, and dependency structure. Phrase structure involves dividing sentences into constituent parts known as syntactic categories, which include both parts of speech and phrasal categories. [8]. This approach generates a tree that conveys phrase structure information and is particularly suited for fixed-order languages such as English. [7]. On the other hand, a dependency relationship is depicted by an arrow from the head to the dependent, along with the name of the relationship [9], [10]. It is preferred for order-free languages such as Urdu which is the national language of Pakistan. Approximately, there are 11 million speakers of Urdu in Pakistan and 300 million plus in the whole world [11] in countries like India, USA, UK, Canada, and USA. With roots in Persian and Arabic, Urdu shares similarities with many South Asian languages, particularly in its lack of capitalization and the absence of distinct small and capital letters. Urdu is comparatively complex as its morphology and syntax structure is a combination of Persian, Sanskrit, English, Turkish, and Arabic [12]. Previously, not much work was done on Urdu Language processing due to little attention from the language engineering community and scarcity of linguistic resources [13].

In this paper, we used MaltParser for Urdu language parsing. MaltParser is a data-driven dependency parsing system that can induce a parsing model from treebank data and apply this model to parse new data [14]. Dependency parsing requires a tagged dataset, prompting us to design a dependency tag-set. We then created this dataset through manual annotation, employing both part-of-speech (POS) tagging and dependency tagging, before conducting experiments with MaltParser. The remainder of this article is structured as follows: Section 2 presents the literature review, while Section 3 discusses the architecture and methodology employed in this study. Section 4 provides details about the dataset, Section 5 discusses the results and finally Section 6 concludes the study and outlines potential future directions.

## 2. LITERATURE REVIEW

Urdu is an under-resourced language, so little work has been done on Urdu. Dependency parsing was explored by [15] using UDT (Urdu dependency Treebank) parsed data and MaltParser, with its default setting. Maltparser can be described as a data-driven parser generator that builds a parser given a Treebank. They removed large sentences from the data, and text including complex ambiguities and punctuation marks was also not considered. Moreover, their tagset was small and no additional features were added to each word. In maltparser, we add features for individual words. The feature model is built based on those features which then helps learners to better learn and classify.

Shift-reduce multi-path strategy-based probabilistic parser was used to analyze Urdu by [16]. There are several limits to a multi-way shift reduction parser for Urdu. It accepts a marked phrase as the entry and cannot comprehend the phrases without POS marking [17]. The analysis of several German dependence parsing systems was performed by [18]. On the same data, four parsers were employed. Two of them were data-driven and two were grammar-driven. MST parser and the Nivre Maltparser data-driven parser seem to function extremely well. Results showed that the learning and parsing models of McDonald and Nivre work very well for German. Nivre's parser has linear time complexity and was the most efficient among other described parsing systems. Both mentioned parsers outperform the results of the rule-based approaches. Another effort for the Urdu parser was made by [19], they parsed Urdu sentences using the Earley parsing algorithm and managed to get an f-score of 87%. The annotation guidelines were divided into 3 sets namely semi-semantic part of speech (SSP), syntactic annotation (SA), and functional annotation.

Another work on dependency parsing of the Hindi language using two parsers MaltParser and MSTParser is done by [20]. Both sentence-level and word-level parsing were done for Hindi. The Treebank used was very small and still under development process. The results showed that MaltParser results were more accurate as compared to MSTParser. Attempts have also been made to create an Urdu treebank for dependency tagging[5]. The evaluation process for annotation was performed by kappa contact. A value of 0.87 was observed and an overall UAS of 74 percent was recorded.

## 3. ARCHITECTURE AND METHODOLOGY

The Urdu dependency parsing system utilized in this study makes use of the MaltParser architecture, receives as input an annotated Urdu dependency treebank (UDT) in CoNLL format. MaltParser, a well-known system for data-driven dependency parsing [21], was trained on UDT to parse the input data, identify dependency relations and label them with head information, as shown below in fig. 1.

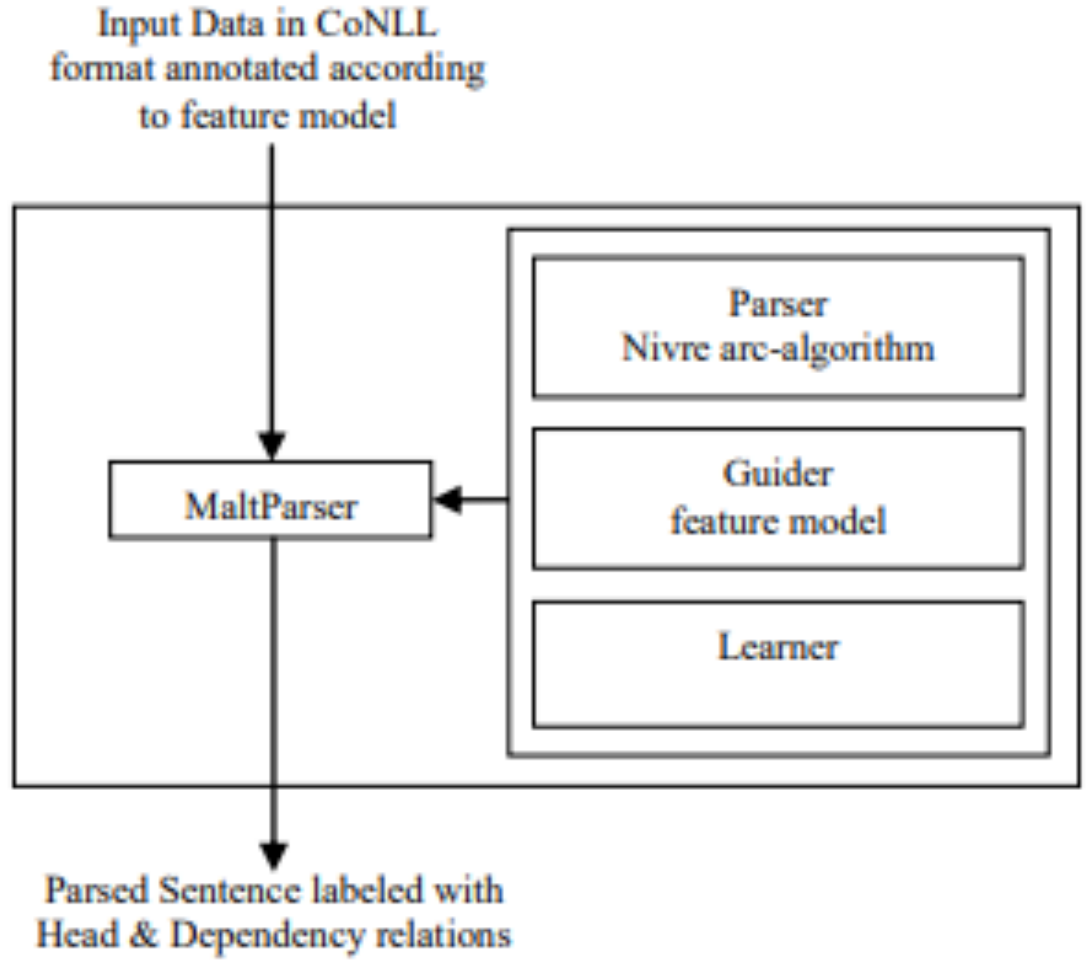

*Figure 1: Urdu Dependency Parser Architecture*

For our experiments, we leveraged MaltParser's flexibility in supporting various parsing algorithms. Different languages often require different parsing strategies, and as such, we generated results using all nine available algorithms along with two classifiers to determine the most suitable approach for Urdu. After thorough evaluation, we found **Nivre's algorithm** to perform best for Urdu. This algorithm, which supports projective dependency structures, has a time complexity of linear order and operates using two primary data structures [22]:

- A stack that holds partially processed token
- An INPUT that holds a list of remaining input tokens

For performance evaluation we used the following metrics: labelled attachment score, unlabeled attachment score, precision, recall and F-score. The LAS measures the percentage of tokens for which both the correct head and the dependency label are identified[23], while the UAS is concerned solely with the accuracy of the head labels. In essence, these metrics are used to measure accuracies at the token level [24]. All test data tokens are taken into consideration and provide equal weighting to each token in the assessment process. The formulas used to calculate LAS and UAS are provided below in equations 1 and 2 respectively:

$$\text{LAS} = \frac{\text{No. of correct head \& dependency labels}}{\text{total tokens}} \quad (1)$$

$$\text{UAS} = \frac{\text{No. of correct head labels}}{\text{total tokens}} \quad (2)$$

These metrics provided a robust evaluation of our system's ability to accurately parse Urdu text.

4. INPUT DATA FORMAT

Since a Treebank is a basic prerequisite for a parser, we created a parsed corpus for Urdu language. Our dataset is comprised of sentences sourced from news articles, encompassing a range of sentence complexities to ensure robust and reliable parsing results—a challenging but essential task. The NU-FAST Treebank [25] was employed which is originally phrase structured. The initial step in our process involved converting the phrase structure into a dependency structure suitable for dependency parsing. In our Urdu Treebank, two layers of tagging are applied.

1. POS (Part of Speech) Tagging
2. Dependency Tagging

Additionally, other linguistic features such as **lemma** and **head information** are incorporated for each word. The POS (part of speech) tag-set in our dataset is based on the work of [26]. However, designing the **dependency tag-set** required a language-specific approach due to the unique characteristics of Urdu. While English, a high-resource language, offers well-established and widely accessible tag-sets[27], these could not be directly applied to Urdu due to significant linguistic differences. Therefore, extensive research and a thorough review of relevant literature were conducted to create a custom dependency tag-set tailored to the morphological and syntactic structure of Urdu and is shown in Table 1.

*Table 1: Dependency tag-set for Urdu*

| S. No | Tag name | Abbreviation |
|---|---|---|
| **1** | Root | Root |
| **2** | Subject | Subj |
| **3** | Direct object | Dobj |
| **4** | Indirect object | Iobj |
| **5** | Noun modifier | Nmod |
| **6** | Verb modifier | Vmod |
| **7** | Numeric modifier | Nummod |
| **8** | Adjective modifier | Adjmod |
| **9** | Adverbial modifier | Advmod |
| **10** | Possession modifier | Poss. |
| **11** | Aspectual auxiliary | Aaux |
| **12** | Tense auxiliary | Taux |
| **13** | Conjunct | Conj |
| **14** | Coordination | Cc |
| **15** | Time period | Tp |
| **16** | Preposition | P |
| **17** | Location | Loc |
| **18** | Quantifier | Q |
| **19** | Reason | R |
| **20** | Negation | NEG |
| **21** | Verb complement | Vcomp |
| **22** | Noun compound modifier | Comp |

The data was then manually annotated into dependency structure, utilizing the dependency tags from table 1. The MaltParser accepts data in a special format called CoNLL format where each line represents a token and characteristic of tokens by tab-separated distance based on which the feature model is created[28]. The selected CoNLL fields that are utilized in this study and contained by the feature model are as follows:

- ID: token counter, initialized by 1 for each new sentence.
- FORM: word form.
- LEMMA: stem word
- CPOSTAG: coarse-grained part-of-speech tag
- POSTAG: Fine-grained POS tag
- FEATS The FEATS field contains a list of morphological features, with a vertical bar (|) as a list separator and with underscore to represent the empty list [29]. All features should be represented as attribute-value pairs, with an equal sign (=) separating the attribute from the value.

- HEAD: head of the current token, which is a value of ID.
- DEPREL: dependency relation of this token with head.

The CoNLL format is helpful for data-driven parsers. As a head/root token of a sentence Zero 0 is used and for any other token the ID of the headword is used as head and the DEPREL value shows the dependency relation between the token and headword. the morphological features used in column FEATS are gender, number and suffix. The sample sentence with POS tags is shown in fig. 2 and its CoNLL format and annotation are demonstrated in Table 2.

1. گجرات<PN>کے<P>سیشن<NN>کورٹس<NN>نے<P>ایک<CA>سال<NN>کے<P>دوران<NN>
بالترتیب<ADV>6231111985<CA>اور<CC>563<CA>مقدمات<NN>کے<P>فیصلے<NN>سنائے<
VB>.<SM>

( S

( KP ( NP ( GP ( PN گجرات ) ( P کے ) ( NP ( NN سیشن
) ( NN کورٹس ) ) ) ( P نے ) )

( NP ( GP ( NP ( CA ایک ) ( NN سال ) ) ( P کے ) )
( NN دوران ) )

( NP ( GP NP ( QP ( ADV بالترتیب ) ( QP ( CA 6231111985 )
( CC اور ) ( CA 563 ) ) ) ( NN مقدمات ) ) ( P کے ) ) ( NN فیصلے )
)

( VP ( VB سنائے ) )

( SM . )

)

*Figure 2:sentence from treebank*

**The Sentence in Roman:** Gujrat ke session courts ne aik saal ke doran bil-tarteeb 6231111985 aur 667 muqadamat ke faisle sunae.

*Table 2: CoNLL format of the above sentence*

| ID | FORM | LEMMA | CPOSTAG | POSTAG | FEATS | HEAD | DEPREL |
|---|---|---|---|---|---|---|---|
| 1 | گجرات | - | PN | PN | \|G=M\|N=S\|Suf=0\| | 4 | Loc |
| 2 | کے | - | P | P | \|G=M\|Suf=0\| | 1 | P |
| 3 | سیشن | - | NN | NN | \|G=M\|N=S\|Suf=0\| | 4 | Comp |
| 4 | کورٹس | - | NN | NN | \|G=M\|N=P\|Suf=0\| | 17 | Subj |
| 5 | نے | - | P | P | \|G=M\|N=S\|Suf=0\| | 4 | P |
| 6 | ایک | - | CA | CA | \|N=S\|Suf=0\| | 7 | Nummod |
| 7 | سال | - | NN | NN | \|G=M\|N=S\|Suf=0\| | 17 | Tp |
| 8 | کے | - | P | P | \|G=M\|Suf=0\| | 7 | P |
| 9 | دوران | - | NN | NN | \|N=S\|Suf=0\| | 7 | Nmod |
| 10 | بالترتیب | - | ADV | ADV | \|N=S\|Suf=0\| | 14 | Advmod |
| 11 | 6231111985 | - | CA | CA | \|N=P\|Suf=0\| | 14 | Nummod |
| 12 | اور | - | CC | CC | \|N=S\|Suf=0\| | 11 | Cc |
| 13 | 563 | - | CA | CA | \|N=P\|Suf=0\| | 11 | Conj |
| 14 | مقدمات | muqadma | NN | NN | \|G=M\|N=P\|Suf=maat\| | 17 | Iobj |
| 15 | کے | - | P | P | \|G=M\|Suf=0\| | 14 | P |
| 16 | فیصلے | faisla | NN | NN | \|G=M\|N=P\|Suf=le\| | 17 | Dobj |
| 17 | سنائے | suna | VB | VB | \|G=M\|N=P\|Suf=ae\| | 0 | Root |

## 5. RESULTS AND DISCUSSION

Table 3 presents the results of these all-parsing algorithms supported by MaltParser using LIBSVM as the learning model, while Table 4 illustrates the results when using Liblinear as the learner. Among all the algorithms, Nivreeager, paired with Liblinear, consistently achieved the highest attachment score, outperforming the other methods.

For evaluation purposes, we used Precision, recall, and F-score values grouped by Dependency relations as shown in Table 5 and Figure 3. High precision value tells that the system is good in ensuring that what is identified is correct.

*Table 3: UAS and LAS using LIBSVM*

| | **UAS** | **LAS** |
|---|---|---|
| **nivreeager** | 82% | 63% |
| **nivrestandard** | 77% | 63% |
| **covnonproj** | 82% | 64% |
| **Covproj** | 80% | 66% |
| **stackproj** | 79% | 64% |
| **stackeager** | 77% | 62% |
| **stacklazy** | 78% | 64% |
| **Planar** | 84% | 63% |
| **2planar** | 83% | 67% |

*Table 4:UAS and LAS using Liblinear*

| | **UAS** | **LAS** |
|---|---|---|
| **nivreeager** | 84% | 70% |
| **nivrestandard** | 77% | 61% |
| **covnonproj** | 82% | 67% |
| **covproj** | 77% | 63% |
| **stackproj** | 80% | 64% |
| **stackeager** | 81% | 64% |
| **stacklazy** | 81% | 67% |
| **planar** | 82% | 72% |
| **2planar** | 84% | 72% |

For example, the dependency tag adjmod has a precision of 80% which means that 80 percent of tag results were relevant. For the same dependency tag, the value for the recall is 0.92 which means that 92 percent of relevant results are returned by the algorithm.

From the results, we can identify which dependency relations yielded higher or lower parsing accuracy and leverage this information to improve the overall system performance by refining specific relations. For instance, the highest accuracy was observed for coordination (CC), negation (NEG), numeric modifier (NUMMOD), and prepositions (P). In the case of negation, the limited set of negation particles in Urdu, often sharing a common stem, enabled the parser to consistently and accurately tag these tokens. A similar explanation applies to numeric modifiers and coordination structures, where the relatively fixed syntactic patterns contributed to the system's high precision and recall. Prepositions, on the other hand, achieved high accuracy because this dependency tag is consistently assigned to words that always belong to the same POS tag; prepositions, reducing ambiguity during parsing. Conversely, the lowest accuracy was observed with location (LOC), indirect object (IOBJ), and noun modifier (NMOD). The possible reason is that these relations often share the same part of speech category, leading to classification challenges.

*Table 5: precision recall and F score grouped by deprel*

| Precision | Recall | F-Score | Deprel |
|---|---|---|---|
| **-** | 0 | - | Aaux |
| **.8** | .923 | .857 | Adjmod |
| **.8** | .571 | .667 | Advmod |
| **1** | 1 | 1 | Cc |
| **.5** | 1 | .667 | Comp |
| **.5** | .333 | .4 | Conj |
| **.556** | .5 | .526 | Dobj |
| **1** | .25 | .4 | Iobj |
| **.333** | .333 | .333 | Loc |
| **1** | 1 | 1 | Neg |
| **.5** | .273 | .353 | Nmod |
| **1** | 1 | 1 | Nummod |
| **1** | 1 | 1 | P |
| **-** | 0 | - | Reason |
| **1** | 1 | 1 | Root |
| **.547** | .667 | .6 | Subj |
| **.75** | 1 | .857 | Taux |
| **.5** | 1 | .667 | Tp |
| **0** | - | - | Vmod |

Table 6 shows both labeled and unlabeled attachment scores grouped by dependency relation and is also depicted in Figure 4. The disparity between labeled and unlabeled scores indicates that system encounters considerable challenges in accurately identifying labels. Dependency relations characterized by unique morphological features and part-of-speech (POS) tags exhibit superior labeled and unlabeled attachment scores. This suggests that the parser demonstrates a high degree of accuracy in identifying these relation labels and their corresponding heads, resulting in minimal differences between labeled attachment scores (LAS) and unlabeled attachment scores (UAS) for relations such as coordinating conjunctions (cc), negation (neg), prepositions (p), and nominal modifiers (nummod). Conversely, dependency relations that share identical POS tags but differ in their dependency tags display lower attachment scores and greater difference in LAS and UAS values as exemplified by tags tp and loc. To overcome this, we used additional morphological features for each token in FEATS column. This increased the performance and reduced the difference a little but still needs improvement. Currently, the morphological features utilized in the FEATS column include gender, number, and suffix. The integration of additional features is expected to yield a more refined and effective feature model.

Finally, Figure 5 presents a comparison between the parsed sentence from the gold standard corpus and the output from the induced parser, illustrating system's performance and pinpoint areas for improvement. From the figure, only one head is incorrectly identified by the system which affects the UAS. However, three relations demonstrated by red labels from parsed sentence shows the relations which are not correctly identified contributing to the lower value for LAS. This observation reinforces the earlier trend where the model exhibited superior UAS compared to LAS.

The statistical tests can be utilized to evaluate the annotation performance of parsers [30]. We utilized the approach of evaluating inter-annotator agreements. Various statistical constants are employed for this purpose. Cohen's kappa is the most utilized constant and has emerged as the de facto benchmark for assessing inter-annotator agreements [31]. It determines how strongly two annotators agree by comparing the probability of the two agreeing by chance with the observed agreement [32]. Mathematically, $\kappa = \frac{p(A)-p(E)}{1-p(E)}$ where $\boldsymbol{\kappa}$ is the kappa value, p(A) is the probability of the actual outcome, and p(E) is the probability of the expected outcome as predicted by chance [33]. The inter-annotator agreement or Kappa value for our system was 0.93 which according to Table 7 is almost perfect agreement. Value could be further increased by enhancing the tagset based on already produced results.

*Table 6: LAS and UAS grouped by dependency relations*

| Deprel | Parser accuracy/LAS metric | Parser accuracy/UAS metric |
|---|---|---|
| **Aaux** | _ | _ |
| **Adjmod** | .667 | .733 |
| **Advmod** | 1.00 | 1.00 |
| **Cc** | 1.00 | 1.00 |
| **Comp** | .500 | .500 |
| **Conj** | .500 | .500 |
| **Dobj** | .556 | .778 |
| **Iobj** | 1.00 | 1.00 |
| **Loc** | .333 | .667 |
| **Neg** | 1.00 | 1.00 |
| **Nmod** | .500 | .500 |
| **Nummod** | .750 | .750 |
| **P** | .947 | .947 |
| **Reason** | _ | _ |
| **Root** | 1.00 | 1.00 |
| **Subj** | .547 | .818 |
| **Taux** | .750 | 1.00 |
| **Tp** | .500 | 1.00 |
| **Vmod** | 000 | .875 |

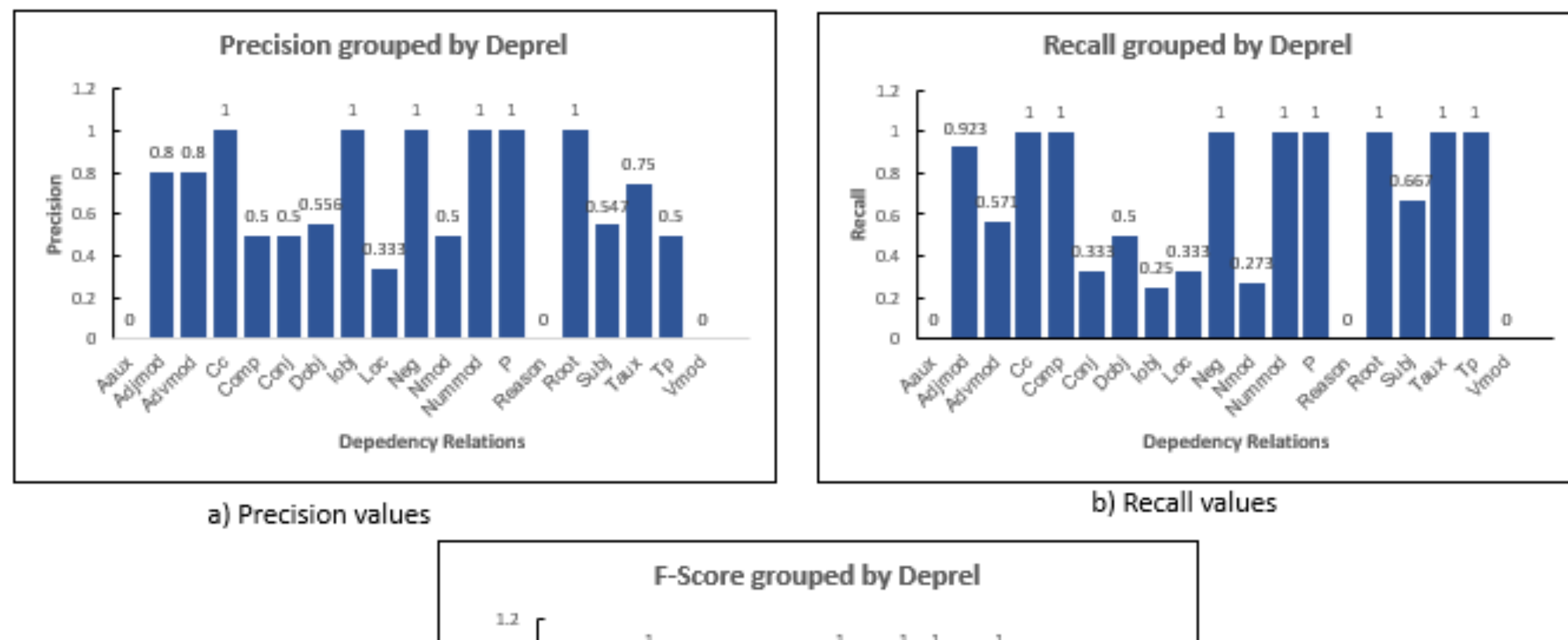

a) Precision values

b) Recall values

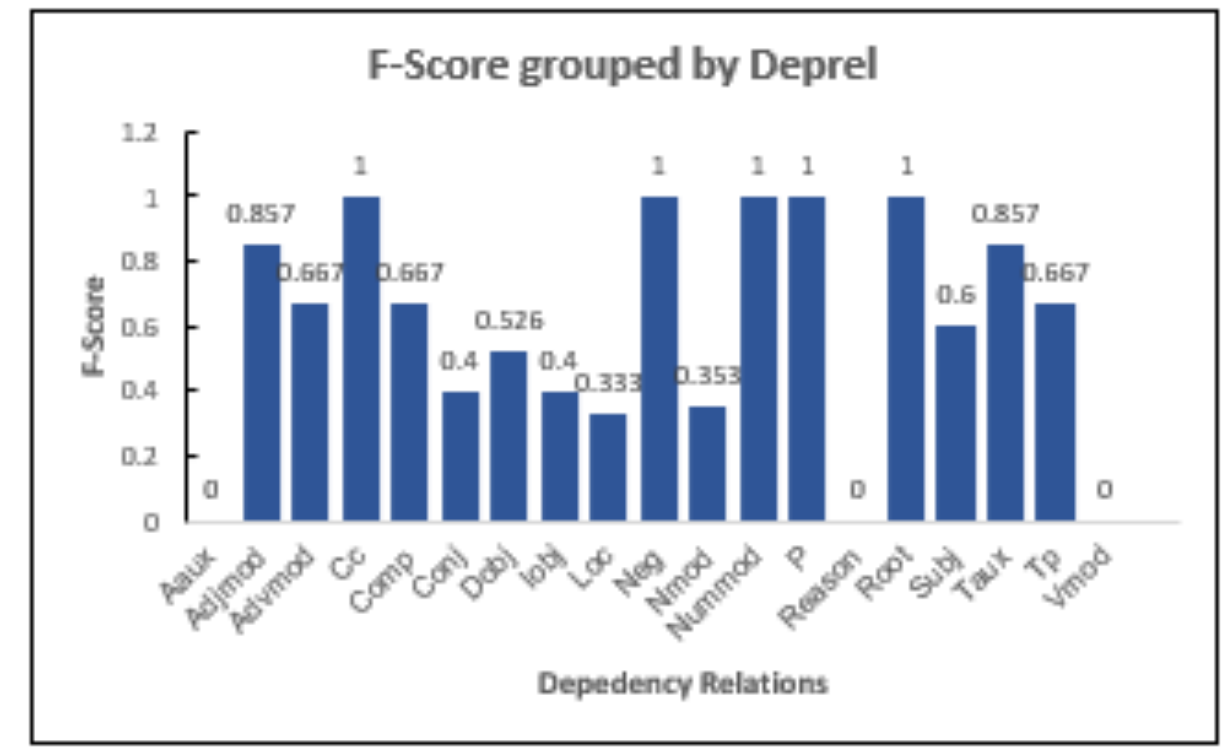

c) F-Score

*Figure 3: evaluation metrics grouped by Deprel, 3a) precision 3b) Recall 3c) F-score.*

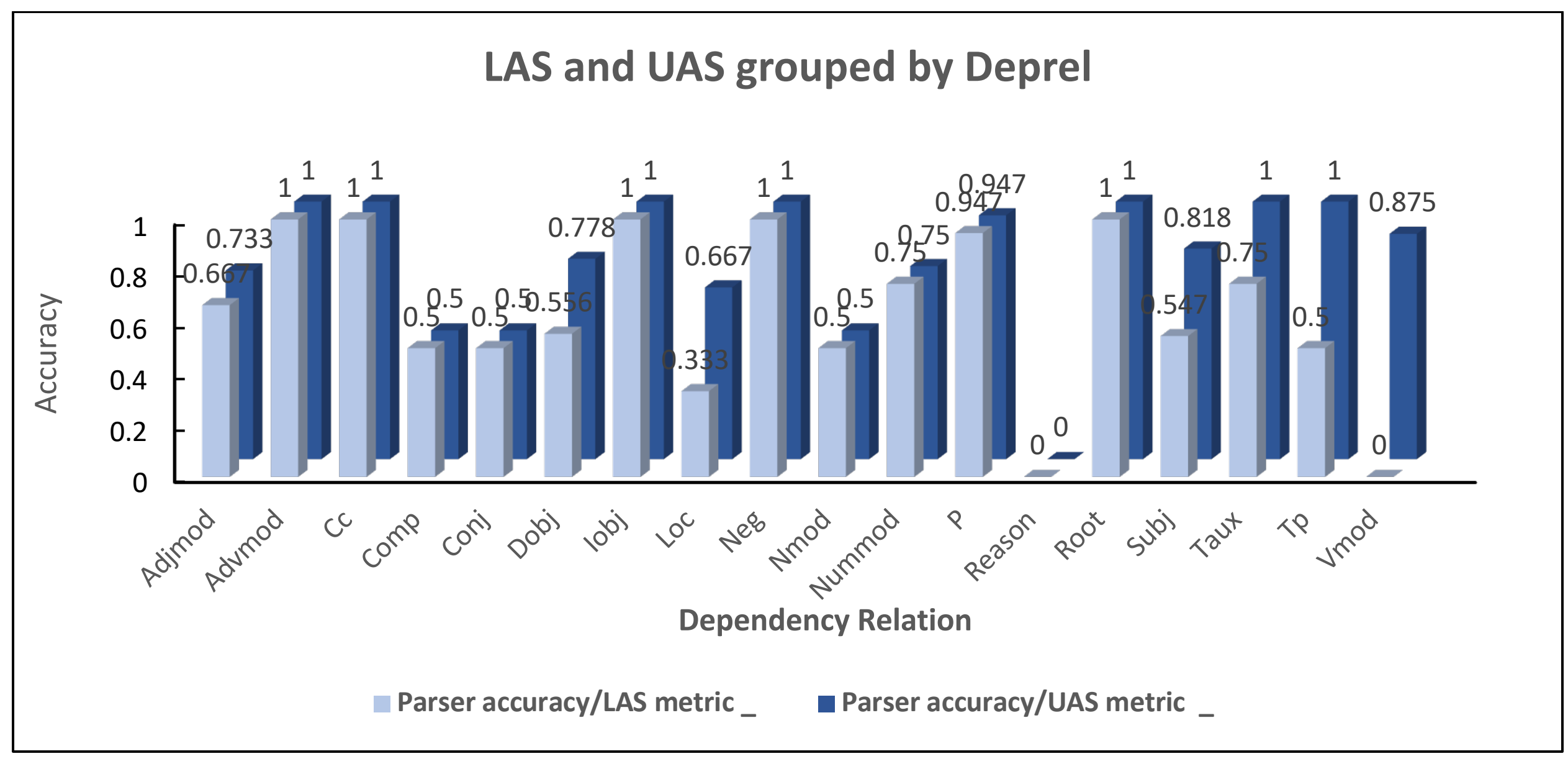

*Figure 4: Parser Accuracy grouped by Deprel.*

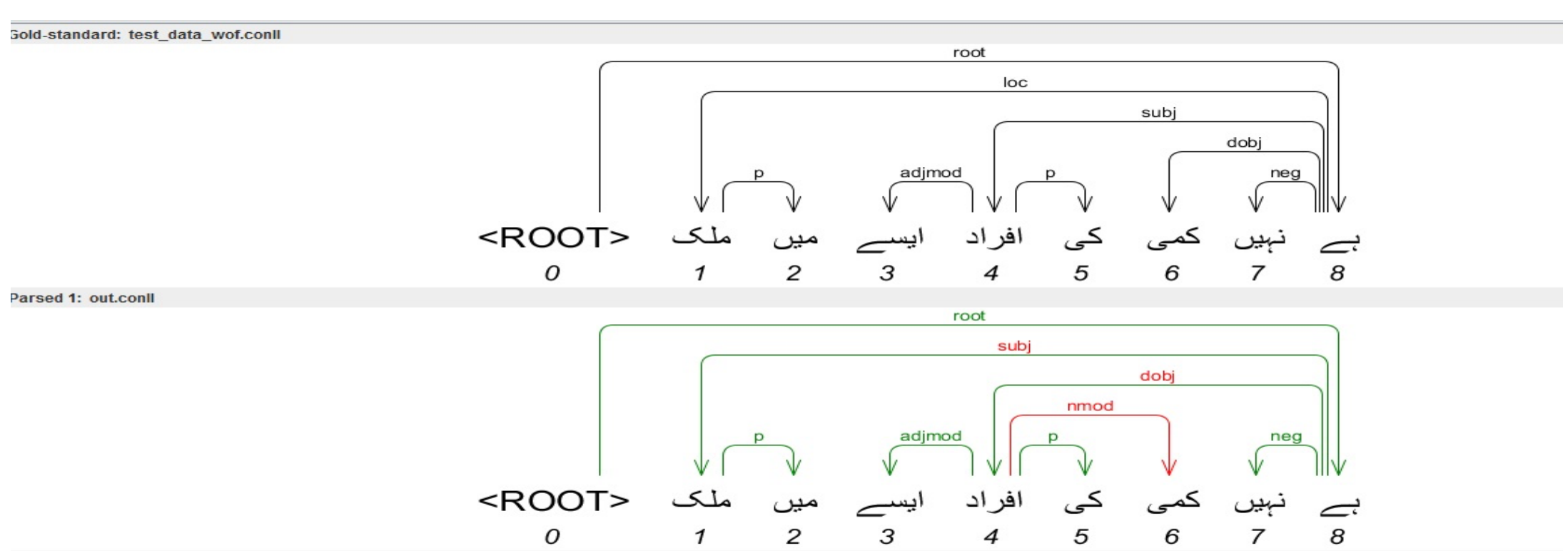

*Figure 5: example of a sentence from the gold corpus and parsed corpus*

*Table 7: Cohen's Kappa value interpretation*

| Value | Agreement |
| --- | --- |
| ≤ 0 | No agreement |
| 0.01-0.20 | None to slight |
| 0.21-0.40 | Fair |
| 0.41-0.60 | Moderate |
| 0.61-0.80 | Substantial |
| 0.81-1.00 | Almost Perfect |

## 6. CONCLUSION AND FUTURE WORK

This study introduces a data-driven, dependency-based approach for parsing Urdu sentences using MaltParser. The dependency dataset and a custom tag-set of 22 labels was created, considering Urdu's complex syntax, morphology, and word order. The system was trained and tested with various feature models and algorithms, with the Nivreeager algorithm delivering the highest accuracy, LAS, and UAS. The results show that MaltParser performs well for Urdu parsing, though accuracy could improve by expanding the Treebank and adding more features. The difference between labeled and unlabeled attachment scores highlights challenges in identifying correct heads. By analyzing precision, recall, and attachment scores, we can identify problematic dependency labels and refine them by incorporating additional morphological features through FEATS to create a more granular feature model. Our dataset's size is limited, so expanding it could improve model training and overall accuracy. Algorithm optimization and parameter fine-tuning in MaltParser, which we have not fully explored, could further enhance performance. FINALLY, incorporating more morphological features into the dataset can boost the parser's performance.

7. REFRENCES

[1] N. CHHILLAR, N. YADAV, and N. JAISWAL, “Parsing: Process Of Analyzing With The Rules Of A Formal Grammar,” *Res. Eng*, pp. 73–79, 2013.

[2] D. Grune and C. J. H. Jacobs, “A programmer-friendly LL (1) parser generator,” *Softw Pract Exp*, vol. 18, no. 1, pp. 29–38, 1988.

[3] G. Thompson, *Introducing functional grammar*. Routledge, 2013.

[4] S. H. Kumhar, M. M. Kirmani, J. Sheetlani, and M. Hassan, “WITHDRAWN: Word Embedding Generation for Urdu Language using Word2vec model,” *Mater Today Proc*, 2021, doi: https://doi.org/10.1016/j.matpr.2020.11.766.

[5] A. Baig, M. U. Rahman, A. S. Shah, and S. Abbasi, “Universal Dependencies for Urdu Noisy Text,” *International Journal*, vol. 10, no. 3, 2021.

[6] J. Bhogal, A. MacFarlane, and P. Smith, “A review of ontology based query expansion,” *Inf Process Manag*, vol. 43, no. 4, pp. 866–886, 2007.

[7] C. Dyer, M. Ballesteros, W. Ling, A. Matthews, and N. A. Smith, “Transition-based dependency parsing with stack long short-term memory,” *arXiv preprint arXiv:1505.08075*, 2015.

[8] N. Xue, F. Xia, F.-D. Chiou, and M. Palmer, “The penn chinese treebank: Phrase structure annotation of a large corpus,” *Nat Lang Eng*, vol. 11, no. 2, pp. 207–238, 2005.

[9] P. Quaresma, V. B. Nogueira, K. Raiyani, and R. Bayot, “Event Extraction and Representation: A Case Study for the Portuguese Language,” *Information*, vol. 10, no. 6, 2019, doi: 10.3390/info10060205.

[10] F. Wu, “Dependency Parsing with Transformed Feature,” *Information*, vol. 8, no. 1, 2017, doi: 10.3390/info8010013.

[11] K. Riaz, “Concept search in Urdu,” in *Proceedings of the 2nd PhD workshop on Information and Knowledge Management*, 2008, pp. 33–40.

[12] F. Adeeba and S. Hussain, “Experiences in building urdu wordnet,” in *Proceedings of the 9th workshop on Asian language resources*, 2011, pp. 31–35.

[13] N. Ahmed, R. Amin, H. Aldabbas, M. Saeed, M. Bilal, and H. Song, “A Novel Approach for Sentiment Analysis of a Low Resource Language Using Deep Learning Models,” *ACM Trans. Asian Low-Resour. Lang. Inf. Process.*, Sep. 2024, doi: 10.1145/3696789.

[14] A. Baig, M. U. Rahman, S. Abrejo, S. Qureshi, S. Tunio, and S. S. Baloch, “Bootstrapping Dependency Treebank of Urdu Noisy Text,” *International Journal*, vol. 9, no. 8, 2021.

[15] W. Ali and S. Hussain, “Urdu dependency parser: a data-driven approach,” in *Proceedings of Conference on Language and Technology (CLT10), SNLP, Lahore, Pakistan*, Citeseer, 2010.

[16] N. Mukhtar, M. A. Khan, F. T. Zuhra, and N. Chiragh, “Implementation of Urdu probabilistic parser,” *International Journal of Computational Linguistics (IJCL)*, vol. 3, no. 1, pp. 12–20, 2012.

[17] A. Weichselbraun and N. Süsstrunk, "Optimizing dependency parsing throughput," in *2015 7th International Joint Conference on Knowledge Discovery, Knowledge Engineering and Knowledge Management (IC3K)*, IEEE, 2015, pp. 511–516.

[18] S. Smirnova, "A Comparison of Dependency Parsers for German," 2006.

[19] Q. Abbas, "Morphologically rich Urdu grammar parsing using Earley algorithm," *Nat Lang Eng*, vol. 22, no. 5, pp. 775–810, 2016.

[20] B. R. Ambati, T. Deoskar, and M. Steedman, "Hindi CCGbank: A CCG treebank from the Hindi dependency treebank," *Lang Resour Eval*, vol. 52, pp. 67–100, 2018.

[21] H. Gharagozlou, J. Mohammadzadeh, A. Bastanfard, and S. S. Ghidary, "Semantic Relation Extraction: A Review of Approaches, Datasets, and Evaluation Methods With Looking at the Methods and Datasets in the Persian Language," *ACM Trans. Asian Low-Resour. Lang. Inf. Process.*, vol. 22, no. 7, Jul. 2023, doi: 10.1145/3592601.

[22] P. Rai and S. Chatterji, "Annotation projection-based dependency parser development for Nepali," *ACM Transactions on Asian and Low-Resource Language Information Processing*, vol. 22, no. 2, pp. 1–19, 2022.

[23] T. Limisiewicz and D. Mareček, "Syntax Representation in Word Embeddings and Neural Networks--A Survey," *arXiv preprint arXiv:2010.01063*, 2020.

[24] T. Nakagawa, "Multilingual dependency parsing using global features," in *Proceedings of the 2007 Joint Conference on Empirical Methods in Natural Language Processing and Computational Natural Language Learning (EMNLP-CoNLL)*, 2007, pp. 952–956.

[25] Q. Abbas, N. Karamat, and S. Niazi, "Development of Tree-bank based probabilistic grammar for Urdu Language," *International Journal of Electrical & Computer Science*, vol. 9, no. 09, pp. 231–235, 2009.

[26] H. Sajjad, "Statistical part of speech tagger for Urdu," *Unpublished MS Thesis, National University of Computer and Emerging Sciences, Lahore, Pakistan*, 2007.

[27] S. Bawa and M. Kumar, "A comprehensive survey on machine translation for English, Hindi and Sanskrit languages," *J Ambient Intell Humaniz Comput*, pp. 1–34, 2021.

[28] S. Pradhan, L. Ramshaw, M. Marcus, M. Palmer, R. Weischedel, and N. Xue, "Conll-2011 shared task: Modeling unrestricted coreference in ontonotes," in *Proceedings of the Fifteenth Conference on Computational Natural Language Learning: Shared Task*, 2011, pp. 1–27.

[29] "scholar (4)".

[30] R. Tsarfaty, J. Nivre, and E. Andersson, "Evaluating dependency parsing: Robust and heuristics-free cross-annotation evaluation," in *Proceedings of the 2011 conference on empirical methods in natural language processing*, 2011, pp. 385–396.

[31] O. Blinova and E. Vilinbakhova, "The Database of Constructions with Lexical Repetitions 'RepLeCon' and Inter-Annotator Agreement," in *International Conference on Internet and Modern Society*, Springer, 2022, pp. 99–114.

[32] M. J. Warrens, “Five ways to look at Cohen’s kappa,” *J Psychol Psychother*, vol. 5, 2015.

[33] S. Vanbelle and A. Albert, “A note on the linearly weighted kappa coefficient for ordinal scales,” *Stat Methodol*, vol. 6, no. 2, pp. 157–163, 2009.